\documentclass[journal]{IEEEtran}

\usepackage{graphicx} 
\usepackage{url}

\usepackage{graphicx}

\usepackage{algorithm} 
\usepackage{algorithmic} 
\usepackage{booktabs}
\usepackage{cite}
\usepackage{xcolor}
\usepackage{multirow}
\usepackage[normalem]{ulem}
\useunder{\uline}{\ul}{}
\usepackage{graphicx}
\usepackage{arydshln}
\usepackage{amsmath}
\usepackage{amssymb} 
\usepackage{orcidlink}

\usepackage{subcaption} 
\begin{document}

\title{Connectivity-Aware Client Selection in Decentralized Federated Learning}

\author{Usevalad Milasheuski \orcidlink{0009-0002-9811-0728}, \IEEEmembership{Member, IEEE}, Monica Nicoli \orcidlink{0000-0001-7104-7015}, \IEEEmembership{Senior Member, IEEE}, and Stefano Savazzi \orcidlink{0000-0002-9865-6512}, \IEEEmembership{Senior Member, IEEE}

\thanks{\textcolor{red}{Received \today;}}
\thanks{Usevalad Milasheuski is with Dipartimento di Elettronica, Informazione e Bioingegneria, Politecnico di Milano, 20133 Milan, Italy, and the Consiglio Nazionale delle Ricerche, IEIIT institute, 20133 Milan, Italy (e-mail:
usevaladmilasheuski@cnr.it).}
\thanks{Monica Nicoli is with Dipartimento di Ingegneria Gestionale, Politecnico di Milano, 20133 Milan, Italy (e-mail: monica.nicoli@polimi.it)}
\thanks{Stefano Savazzi is with the Consiglio Nazionale delle Ricerche, IEIIT institute, 20133 Milan, Italy (e-mail: stefano.savazzi@cnr.it).}}

\markboth{Journal of \LaTeX\ Class Files, Vol. 14, No. 8, August 2015}
{Shell \MakeLowercase{\textit{et al.}}: Bare Demo of IEEEtran.cls for IEEE Journals}
\maketitle

\begin{abstract}
Decentralized federated learning (DFL) has emerged as a scalable alternative to centralized federated learning, removing the single point of failure and communication bottleneck at the server. However, most existing decentralized approaches rely on full-neighborhood broadcasting, where each client transmits its model update to all of its neighbors in each communication round. While effective, this results in significant communication overhead and high energy consumption. This paper applies spectral graph tools as lightweight client selection methods that leverage the graph topology by analyzing the spectral properties of the graph Laplacian over a fixed communication graph. Specifically, we exploit eigenvector embeddings and heat/diffusion kernel similarities to guide each client in selecting only a subset of structurally relevant neighbors, i.e., those that are most beneficial for the convergence of the federation process. This topology aware randomization matches or outperforms uniform neighborhood subsampling and approaches the accuracy of full broadcasting. The largest gains, about $3$ percentage points, emerge under strong data heterogeneity on larger, structurally rich graphs, while the total number of transmitted messages is reduced by $50\%$ compared to naive broadcasting.
\end{abstract}


\IEEEpeerreviewmaketitle

\section{Introduction}
\label{sec:intro}

\IEEEPARstart{F}{ederated} Learning (FL) is a distributed paradigm that enables collaborative training of a global model across decentralized edge devices without sharing raw data \cite{FL}. Unlike centralized training, where data are aggregated on a single server, FL performs on-device training and transmits only model updates. This preserves privacy, reduces communication costs, and makes FL a promising solution for diverse applications \cite{healthcare, drones}. While FL provides significant privacy benefits, its reliance on a central server for model aggregation introduces a single point of failure. The entire training process is susceptible to server-side issues, including hardware failures, network outages, or malicious attacks, which can halt the process and compromise the global model's integrity \cite{attack1, attack2}. To overcome this vulnerability, a paradigm known as Decentralized Federated Learning (DFL) has emerged. DFL eliminates the central server, instead leveraging a peer-to-peer network where devices communicate directly with their neighbors to share and aggregate model updates. This distributed architecture not only removes the single point of failure but also improves scalability and resilience, as the system can continue to operate even if some devices fail or become disconnected \cite{decSGDKOLOS, DFLsurvey, DecSGDTop}.

\begin{figure} 
\centerline{\includegraphics[width=0.48\textwidth]{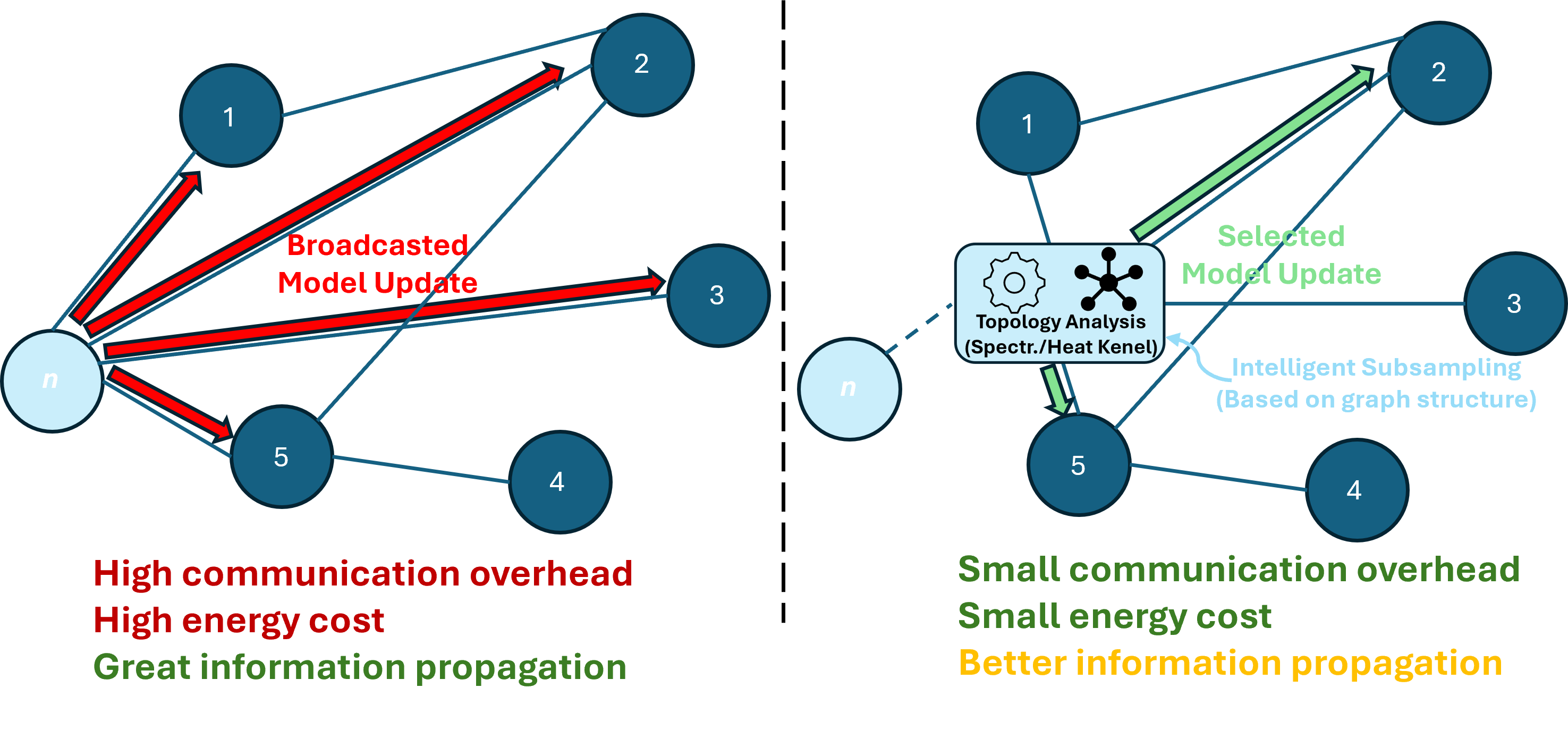}}
    \caption{\textbf{Standard Broadcasting} (Left): A client sends updates to its entire neighborhood. This approach leads to high communication overhead and rapid energy depletion. \textbf{Topology-Aware Selection} (Right): Clients use spectral properties to identify and message only structurally significant neighbors. This strategic subsampling reduces costs while maintaining effective information flow compared to random selection.}
\label{fig:cont_learn_fig}
\vspace{-0.4cm}
\end{figure}

Most DFL studies assume a complete communication graph, where each client transmits its updates to all others \cite{DSGD, healthcare, decOPT, iot}. This all-to-all scheme simplifies convergence analysis but incurs high communication overhead, often prohibitive in bandwidth- or energy-constrained systems \cite{decOPT}. To overcome this, more practical approaches design sparse topologies or intelligent client selection \cite{graph_select1, graph_select2}, typically constructing the graph by (i) thresholding a similarity measure over a complete graph \cite{smart_samp}, (ii) selecting neighbors by model or gradient similarity \cite{select_grad_1}, or (iii) accounting for divergences in local data distributions \cite{select_data_2}. However, these methods ignore the fixed communication topology, which restricts connections between certain pairs of nodes.

This letter proposes a \textit{connectivity-aware neighbor selection mechanism} that mitigates the communication and energy inefficiencies of naive neighborhood broadcasting in DFL. Each client selects a subset of its neighbors according to Laplacian-based similarity measures computed from the client adjacency graph. Our decentralized client selection framework relies exclusively on the communication topology and does not require access to local data distributions or model parameter dynamics, making it more energy- and communication-efficient than full broadcasting. Within this unified framework, we investigate two alternative ways of deriving topology-aware similarities from the graph Laplacian, which are then used to guide the same neighbor selection process:
\begin{itemize}
    \item \textbf{Spectral Embedding}: clients are represented in a low-dimensional space induced by selected Laplacian eigenvectors, and neighbors are chosen based on proximity in this spectral embedding.
    \item \textbf{Heat Kernel Similarity}: client similarity is defined through diffusion dynamics over the graph, where neighbors with stronger heat kernel interactions are preferentially selected.
\end{itemize}
The paper is organized as follows. We review related work in Section \ref{sec:relwork}, discuss the methodology behind the topology-aware client selection in Section \ref{sec:methodology}. Section \ref{sec:exper} presents experimental results that validate our approach, and we summarize our findings in Section \ref{sec:concl}.
\vspace{-0.3cm}
\section{Related Work and Contributions}
\label{sec:relwork}

DFL's application domains---drones \cite{drones}, IoT \cite{iot}, and healthcare \cite{healthcare}---demand efficient communication, yet decentralized methods often assume complete topologies and broadcasting \cite{DSGD}, which can make communication inefficient despite strong convergence guarantees. For example, the centralized FedAvg algorithm \cite{FL} randomly selects a subset of clients each round to address this issue. Some methods use importance sampling \cite{imp_samp1, imp_samp2}, prioritizing clients with large gradient norms, while others measure the gradient or model alignment between models \cite{client_GRAD_sampling, client_GRAD_sampling2} to effectively subsample the neighborhood. In decentralized setups, a new graph is often constructed over the complete graph topology by specifying the adjacency matrix or learning implicit similarities from data variability, update norms, or validation accuracies \cite{graph_select1, graph_select2}. While effective under heterogeneous data, decentralized methods ignore predefined client topologies, causing bottlenecks and inefficient information flow. In contrast, graph-theoretic techniques evaluate nodes using structural metrics (e.g., degree, centrality) or random-walk methods \cite{deepwalk, node2vec}.

A particularly powerful and rich class of methods for graph representation involves spectral graph theory \cite{graph_theory}. Unlike simple ad-hoc methods, spectral methods analyze the graph's properties through the eigenvalues and eigenvectors of the Laplacian matrix. The eigenvectors of the graph Laplacian encode a unique representation of each node's position and its influence on the network's overall connectivity and information flow. This leads to a different approach for client selection, where nodes can be prioritized based on their spectral properties to guide the aggregation process, reducing communication while keeping accuracy close to full broadcasting. Such spectral techniques are used for multiple applications, including graph partitioning, node clustering \cite{graph_theory} and positional encoding for transformers \cite{pos_emb}. Spectral tools have also been applied to DFL. For instance, SCDFL~\cite{scdfl} forms a topology via spectral clustering, while Laplacian sampling~\cite{laplacian_sampling} and graph sparsification~\cite{graph_sparsify} prune a global mixing matrix to save energy. Additionally, probabilistic schemes~\cite{proba_comm} skip links under heterogeneity; the role of topology in decentralized convergence is itself an active question~\cite{topo_role}. However, these existing approaches typically construct new topologies, alter the underlying network, or rely on global matrix operations, leaving open the question of how to efficiently select clients over a strictly fixed, predefined communication graph.

The paper investigates a topology based neighbor selection mechanism for DFL. Specifically, for a \emph{fixed} graph, purely topological Laplacian similarities are computed once on the public topology and then used at each round to select the $\beta$-fraction of \emph{existing} neighbors to contact, without altering the topology or exchanging data/gradients.

\vspace{-0.3cm}
\section{Methodology}
\label{sec:methodology}

This section introduces the proposed methods for neighbor selection based on the spectral properties of the topology graph of decentralized clients. The goal is to efficiently select the most effective nodes to federate instead of the entire neighborhood to reduce the communication overhead and improve model convergence and generalization.

\subsection{Problem Setup} 
We consider an undirected  graph $\mathcal{G}=(\mathcal{V},\mathcal{E})$ where $\mathcal{V} = \{1,2,\dots,N\}$ is the set of $N$ nodes/clients and $\mathcal{E} \subseteq \mathcal{V} \times \mathcal{V}$ is the set of communication links. The adjacency matrix $\mathbf{A} \in \{0,1\}^{N \times N}$ is defined as
\begin{equation}
    A_{ij} =
        \begin{cases}
            1, & \text{if } (i,j) \in \mathcal{E} \text{ and } i \neq j, \\
            0, & \text{otherwise.}
        \end{cases}
\end{equation}
The unnormalized Laplacian of a graph is defined as \cite{graph_theory}:
\begin{equation}
    \mathbf{L} = \mathbf{D} - \mathbf{A},
\end{equation}
where $\mathbf{D}$ is a diagonal degree matrix with entries $D_{ii} = \sum_{j=1}^N A_{ij}, \quad \forall i \in \{1, \dots, N\}$. For spectral analysis, a normalized version of the Laplacian is commonly used:
\begin{equation}
    \mathbf{L}_{\text{sym}} = \mathbf{D}^{-1/2}\mathbf{L}\mathbf{D}^{-1/2}.
\end{equation}

Both Laplacians $\mathbf{L}$ and $\mathbf{L}_{\text{sym}}$ are symmetric and positive semi-definite, and $\mathbf{L}_{\text{sym}}$ admits the eigenvalue decomposition
\begin{equation}
\mathbf{L}_{\text{sym}} = \mathbf{U} \mathbf{\Lambda} \mathbf{U}^\top,
\end{equation}
where $\mathbf{U} = [\mathbf{u}_1, \mathbf{u}_2, \dots, \mathbf{u}_N] \in \mathbb{R}^{N \times N}$ is an orthonormal matrix of eigenvectors,
and $\mathbf{\Lambda} = \mathrm{diag}(\lambda_1, \lambda_2, \dots, \lambda_N)$ contains the eigenvalues 
ordered as $0 = \lambda_1 \leq \lambda_2 \leq \cdots \leq \lambda_N$.

The Laplacian spectrum reveals structural and dynamical properties of the network. The eigenvalues act as graph \textit{frequencies}: the smallest nonzero one reflects how well clients are linked, with larger values enabling faster consensus \cite{graph_theory}. The eigenvectors are graph harmonics; smooth ones (small eigenvalues) vary slowly across connected clients, while higher modes capture local fluctuations. In what follows, we exploit these properties for neighborhood subsampling.
\vspace{-0.3cm}
\subsection{Spectral Embedding} 
Laplacian eigenvectors form the basis of spectral embeddings, mapping each client into a low-dimensional Euclidean space. Let $\mathbf{U}_k \in \mathbb{R}^{N \times k}$ collect the $k$ eigenvectors $\mathbf{u}_i=[u_{i,n}]_{n=1}^{N}$, $i=2, ...,k+1$ corresponding to $k$ smallest nonzero eigenvalues, the embedding $\mathbf{e}_n$ of node $n$ is given by the normalized $n$-th \textit{row} of $\mathbf{U}_k$:
\begin{equation} 
    \mathbf{e}_n = [\tilde{u}_{2,n}, \tilde{u}_{3,n}, \dots, \tilde{u}_{k+1,n}], 
\label{eq:embeddings}
\end{equation} 
where $\tilde{u}_{i,n} = u_{i,n}/(\sum_{j=2}^{k+1} u_{j,n}^2 )^{1/2}$ normalizes each embedding to unit norm, so that (\ref{eq:cos_sim}) is a cosine similarity and per-node magnitude does not bias selection.
By selecting the $k$ smallest non-zero eigenvectors, the embedding from (\ref{eq:embeddings}) sets a sufficient representation of each node. Low-frequency eigenvectors are well suited to our objective of promoting collaboration over the graph: smooth modes group structurally proximal clients and, when heterogeneity is topology-correlated (Sec.~\ref{sec:topo_exp}), also track distributional similarity. The full spectrum is needed mainly on heterophilous graphs, where adjacency anti-correlates with node attributes~\cite{heterophily_lpe}; $k$ can be increased to retain higher modes. Taking inspiration from node embedding techniques \cite{deepwalk, node2vec}, we compare nodes by the similarity of their embeddings: 
\begin{equation} \label{eq:cos_sim}
    S^{spectr}(n_1, n_2) = \sum_{i=2}^{k+1}  \tilde{u}_{i,n_1} \cdot \tilde{u}_{i,n_2}.
\end{equation}
A high similarity indicates that two clients occupy similar \textit{directions} in spectral space, sharing comparable neighborhoods and roles in information diffusion.
\vspace{-0.3cm}
\subsection{Heat Kernel}
The spectral properties of the Laplacian also allow the construction of diffusion-based similarity measures. A central example is the heat kernel, which models the propagation of information across the communication graph in analogy with heat diffusion over a surface \cite{heat_kern}. Formally, the heat kernel at diffusion time $t > 0$ is defined as
\begin{equation} \label{eq:dif_ker}
    \mathbf{K}_t = \exp(-t\mathbf{L}) = \sum_{n=1}^N e^{-t\lambda_n}\mathbf{u}_n \mathbf{u}_n^T, 
\end{equation}
where small values of $t$ emphasize local structure, highlighting immediate neighborhoods, while larger values of $t$ reflect long-range connectivity and global mixing behavior. Each entry $K_{n_1 n_2}(t)$ of $\mathbf{K}_t$ captures the amount of \textit{heat} transferred from node $n_1$ to node $n_2$ after time $t$. As an alternative to Eq. \ref{eq:cos_sim}, heat kernel Eq. \ref{eq:dif_ker}  can be directly used as a measure of similarity between nodes:
\begin{equation}\label{eq:diff_dist}
    S^{heat}(n_1, n_2) =  (K_{n_1 n_1}(t) + K_{n_2 n_2}(t) - 2K_{n_1 n_2}(t))^{-1},
\end{equation}
so that nodes exchanging more heat, i.e., better diffusively connected, are more likely to be selected. In practice, $\mathbf{K}_t$ is formed from the same $k$ low-frequency Laplacian modes used as embeddings, so both selectors share the dimension $k$ \cite{heat_kern}.
\vspace{-0.3cm}
\subsection{Decentralized Federated Learning}
A set of clients is defined over the graph $G$. Each client $n$ holds local data $\mathcal{D}^n={\mathbf{x}^n_i, y^n_i}_{i=1}^{|\mathcal{D}^n|}$, where $\mathbf{x}^n_i$ and $y^n_i$ denote its local features and labels. Generally, the data is not Independent and Identically Distributed (non-IID) among the clients, leading to highly heterogeneous setups. The goal is to collaboratively find a set of model parameters $\boldsymbol{w}$ that will minimize the average loss over the nodes
\begin{equation} \label{eq:optim}
    \min_{\boldsymbol{w} \in \mathbb{R}^d} L(\boldsymbol{w})=\min_{\boldsymbol{w} \in \mathbb{R}^d} \frac{1}{N}\sum_{n=1}^{N}L_n(\boldsymbol{w}),
\end{equation}
where  $L_n(\boldsymbol{w})=\frac{1}{|\mathcal{D}^n|}\sum_{(\mathbf{x}^n_i, y^n_i) \in \mathcal{D}^n}l(\boldsymbol{w}, (\mathbf{x}^n_i, y^n_i))$. 

Given a node $n$, we define a subsampled neighborhood $\mathcal{N}_{\beta}(n) \subseteq \mathcal{N}(n)$, where the parameter $\beta \in (0,1]$ controls the subsampling ratio. In classical decentralized optimization frameworks that minimize the objective~(\ref{eq:optim}), full neighborhood communication is adopted, which corresponds to the special case $\beta = 1$.  

At each communication round $t$, each node: \textbf{1)} trains the model on its local dataset; \textbf{2)} forwards its model parameters to selected neighbors; and \textbf{3)} aggregates the received models:
\begin{equation} \label{eq:aggreg}
    \boldsymbol{w}_n^{t+1} = \rho \boldsymbol{w}_n^{t} + (1 - \rho)\frac{1}{|\mathcal{N}_{\beta}(n)|}\sum_{m \in \mathcal{N}_{\beta}(n)} \boldsymbol{w}_m^{t},
\end{equation}
where $|\mathcal{N}_{\beta}(n)| = \left\lceil \beta\,|\mathcal{N}(n)| \right\rceil$. The subset $\mathcal{N}_{\beta}(n)$ is obtained by sampling $|\mathcal{N}_{\beta}(n)|$ distinct nodes from $\mathcal{N}(n)$ according to a Laplacian-based similarity measure. Specifically, the probability of selecting neighbor $n_2 \in \mathcal{N}(n_1)$ is defined as
\begin{equation} \label{eq:proba}
    p(n_1, n_2) = \frac{\exp\!\left(S(n_1, n_2)\right)}{\sum_{n \in \mathcal{N}(n_1)} \exp\!\left(S(n_1, n)\right)},
\end{equation}
where the similarity function $S(n_1,n_2)$ is computed either via the spectral embedding $S^{spectr}(n_1,n_2)$ in~(\ref{eq:cos_sim}) or via the heat kernel diffusion $S^{heat}(n_1,n_2)$ in~(\ref{eq:diff_dist}).

\begin{figure}[t]
    \centering
    \begin{subfigure}[b]{0.49\linewidth}
        \centering
        \includegraphics[width=\linewidth]{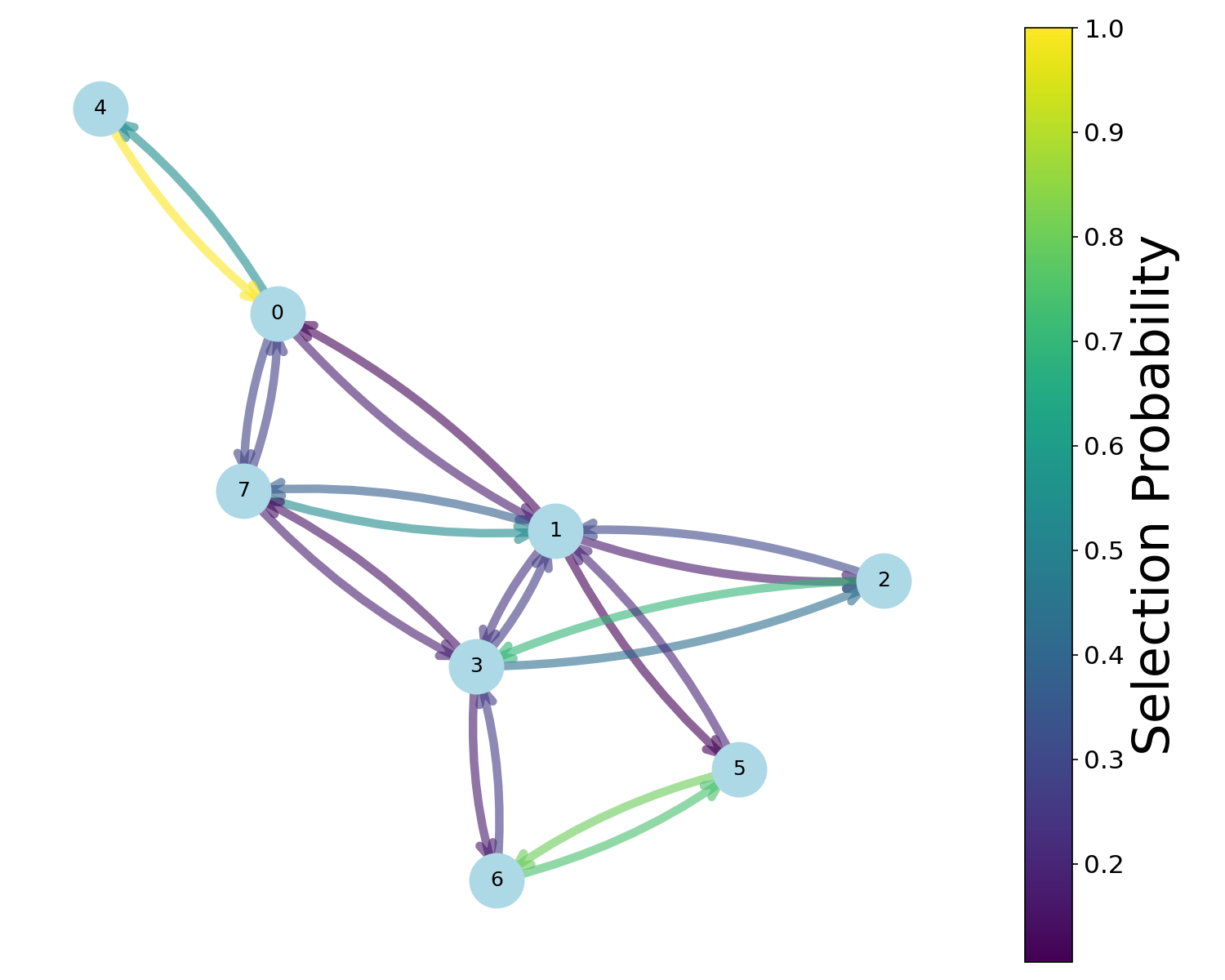}
        \caption{Spectral Embedding}
        \label{fig:res3}
    \end{subfigure}
    \hfill
    \begin{subfigure}[b]{0.49\linewidth}
        \centering
        \includegraphics[width=\linewidth]{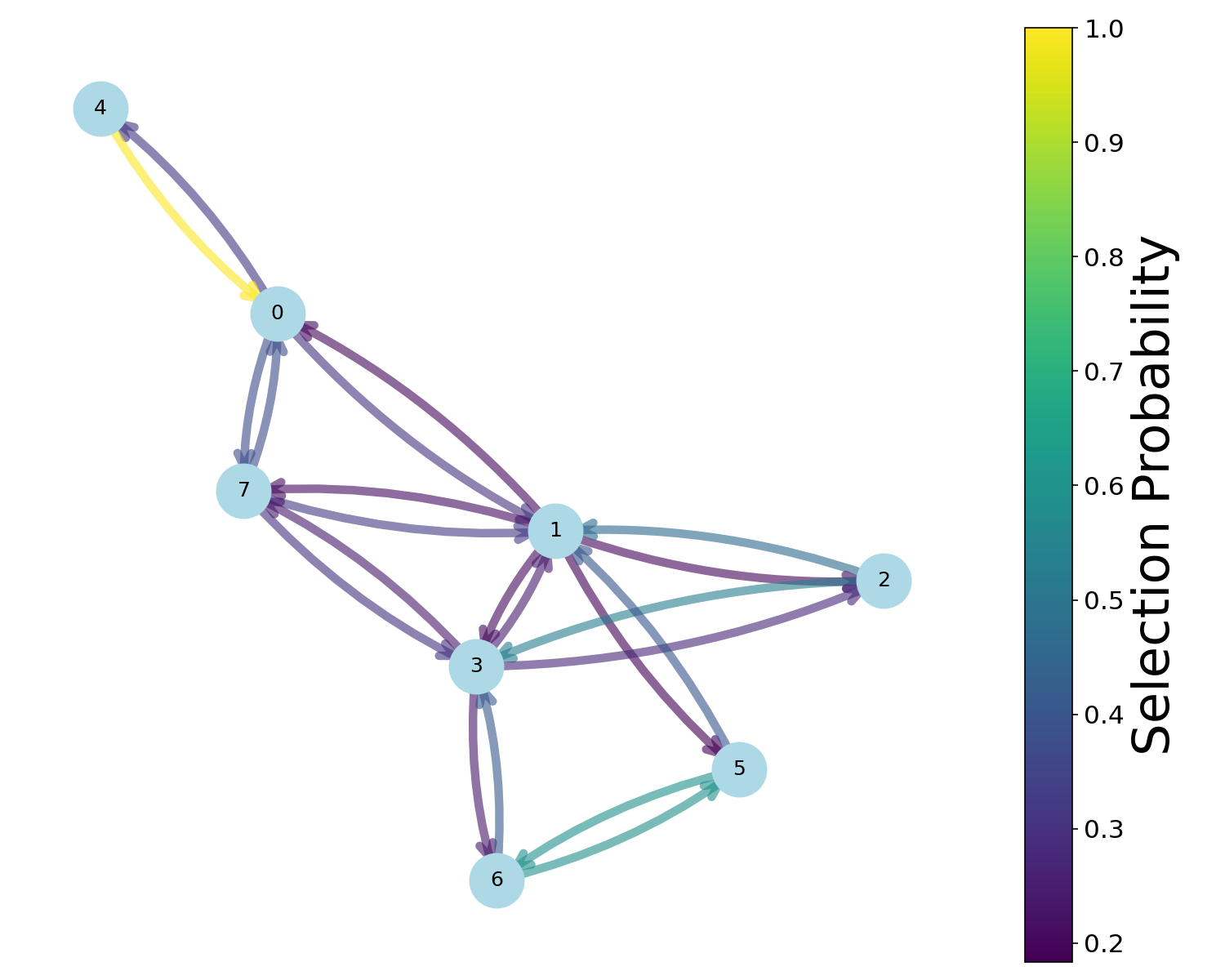}
        \caption{Heat Kernel}
        \label{fig:res4}
    \end{subfigure}
    \caption{Selection probabilities for \textcolor{blue}{\textit{Spectral Embedding}} (left) and \textit{Heat Kernel} methods on a \textit{small} graph.}
    \label{fig:res}
    \vspace{-0.4cm}
\end{figure}

For the graph in Figure \ref{fig:res}, the transmission probabilities between connected nodes are shown using a color scale, to highlight how likely information is to be transmitted along each edge.


\vspace{-0.3cm}
\subsection{Privacy and Adaptability Considerations}
The connectivity graph is assumed to be given as an input to the proposed method. In the considered settings, the topology is determined by physical communication range or routing constraints, and can be either publicly available or inferred through consensus-driven protocols \cite{learn_graph} without relying on a central server or sharing local data or model updates. In practice, this is a \emph{one-time} $\mathcal{O}(N^3)$ precomputation (or $\mathcal{O}(k|\mathcal{E}|)$ with iterative solvers), after which selection is $\mathcal{O}(|\mathcal{N}(n)|)$ per round---cheaper than gradient-based selectors~\cite{client_GRAD_sampling, node_select_niid} whose cost scales with model dimension $d \gg N$; learning-aware methods~\cite{byzantine_dsgd, node_select_niid} are complementary and can be layered on the $\beta$-subsampled neighborhood.

Another important aspect of our method might be related to dynamic graphs (e.g., swarm learning \cite{swarm}). While they pose distinct challenges, our framework can adapt using iterative methods like Locally Optimal Block Preconditioned Conjugate Gradient (LOBPCG) \cite{learn_lapl} to quickly learn new eigenvectors based on prior states.

\vspace{-0.4 cm}
\hspace{-0.3cm}
\section{Experiments}
\label{sec:exper}

In this section we evaluate the proposed tool on three graphs of increasing size: a synthetic $8$-node Erdős-Rényi graph \cite{random_graph} (\textbf{small}, edge prob.\ $0.4$), the $32$-node Davis Southern women network \cite{south_graph} (\textbf{medium}), and the $77$-node Les Miserables co-appearance network \cite{miserable_graph} (\textbf{large}). We use FMNIST and CIFAR10 under a symmetric Dirichlet split with $\alpha=0.3$ \cite{dir}, with the models of \cite{FL} (FMNIST replacing MNIST). Training uses SGD for $3$ local epochs (lr $0.01$, batch $64$, $\rho=0.9$), $100$ rounds, averaged over $3$ runs. For selection we set $\beta=0.5$, with $k{=}3,6,12$ for the \textit{small}/\textit{medium}/\textit{large} graphs and heat-kernel time $t{=}0.01$ (\textit{small}) or $0.05$ (larger).

In what follows, we compare the following setups: 1) \textbf{No-federation}, independent local training; 2) \textbf{Broadcasting} to all neighbors; 3) \textbf{Random Sampling} of a $\beta$-fraction of neighbors; 4) \textbf{Gradient Similarity}, using the angle between normalized node/neighbor gradients \cite{client_GRAD_sampling}; 5) \textbf{Spectral Embedding}, via (\ref{eq:cos_sim}) and (\ref{eq:proba}); 6) \textbf{Heat Kernel}, via (\ref{eq:dif_ker}) and (\ref{eq:proba}); and 7) \textbf{Data Similarity}, a data-aware baseline that samples neighbors with similar label distributions (low KL divergence)~\cite{feddkw}.

\vspace{-0.3cm}
\subsection{Performance Analysis} 
\label{sec:res}
Table~\ref{tab:res_table} shows clear patterns across datasets and graph sizes: \textit{no-federation} is the lower bound and \textit{broadcasting} the upper one at maximal communication, while our topology-aware subsampling sits in between, consistently beating random at lower communication. On FMNIST, \textit{spectral embedding} is on par with the gradient-similarity baseline, despite using \emph{no} gradient information, which is the practically relevant comparison since our selector is data- and gradient-free. On the harder CIFAR10 task the differences shrink but persist, with the \textit{heat kernel} best on the small and large graphs and \textit{spectral embedding} best on the medium one. The data-aware \textit{Data Sim.}\ baseline is weakest throughout, below random in every cell: naively aggregating with label-similar neighbors is counter-productive. These $\alpha{=}0.3$ margins are modest but, as shown next, widen markedly under stronger heterogeneity on the larger graph.
\begin{table}[]
\centering
\begin{tabular}{|c|c|clcll|}
\hline
\multirow{2}{*}{Dataset} & \multirow{2}{*}{Selection}  & \multicolumn{5}{c|}{Topology}                                                                           \\ \cline{3-7} 
                         &                             & \multicolumn{2}{c|}{Small }             & \multicolumn{2}{c|}{Medium }            & Large \\ \hline
\multirow{7}{*}{FMNIST}  & \textcolor{gray}{Broadcast} & \multicolumn{2}{c}{\textcolor{gray}{86.36}} & \multicolumn{2}{c}{\textcolor{gray}{84.13}} & \textcolor{gray}{76.84}            \\
                         & \textcolor{gray}{No-Fed.}   & \multicolumn{2}{c}{\textcolor{gray}{65.65}} & \multicolumn{2}{c}{\textcolor{gray}{61.49}} & \textcolor{gray}{61.30}            \\
                         & Random                      & \multicolumn{2}{c}{81.01}                   & \multicolumn{2}{c}{80.84}                   &  74.17           \\
                         & Data Sim. & \multicolumn{2}{c}{76.23} & \multicolumn{2}{c}{77.68} & 72.36           \\
                         & Grad. Sim.                  & \multicolumn{2}{c}{\textbf{82.53}}          & \multicolumn{2}{c}{81.36}                   &  \textbf{75.21}          \\
                         & Spectr. Emb.                & \multicolumn{2}{c}{\underline{82.44}}        & \multicolumn{2}{c}{\textbf{82.15}}         &  \underline{74.83}           \\
                         & Heat Kern.                  & \multicolumn{2}{c}{81.31}                   & \multicolumn{2}{c}{\underline{81.86}}       &  74.48           \\ \hline
\multirow{7}{*}{CIFAR10} & \textcolor{gray}{Broadcast} & \multicolumn{2}{c}{\textcolor{gray}{62.02}} & \multicolumn{2}{c}{\textcolor{gray}{54.53}} & \textcolor{gray}{43.15}           \\
                         & \textcolor{gray}{No Fed.}   & \multicolumn{2}{c}{\textcolor{gray}{40.68}} & \multicolumn{2}{c}{\textcolor{gray}{31.53}} & \textcolor{gray}{29.74}            \\
                         & Random                      & \multicolumn{2}{c}{57.31}                   & \multicolumn{2}{c}{48.87}                   & 39.76            \\
                         & Data Sim. & \multicolumn{2}{c}{56.14} & \multicolumn{2}{c}{45.82} & 39.01           \\
                         & Grad. Sim.                  & \multicolumn{2}{c}{57.50}                    & \multicolumn{2}{c}{48.64}                   & 40.07           \\
                         & Spectr. Emb.                & \multicolumn{2}{c}{\underline{57.74}}         & \multicolumn{2}{c}{\textbf{50.13}}        & \underline{40.29}            \\
                         & Heat Kern.                  & \multicolumn{2}{c}{\textbf{58.05}}          & \multicolumn{2}{c}{\underline{49.89}}       & \textbf{40.39}           \\ \hline
\end{tabular}
\caption{Average model accuracy across datasets and client topologies. The \textbf{bold values} indicate the best subsampling method (excluding Broadcast); \underline{underlined} is the second best.}
\label{tab:res_table}
\end{table}

\vspace{-0.3 cm}
\subsection{Topology-correlated heterogeneity and sensitivity}
\label{sec:topo_exp}
The Dirichlet partition in Table~\ref{tab:res_table} assigns data independently of the graph. Two aspects remain open: whether the topology encodes \emph{learning-relevant} structure, and under which conditions the selection rule pays off. Two settings are proposed in the following. The first is a \textbf{topology-correlated} partition (\textit{Topo}) on the \textit{medium} graph, with labels varying smoothly along the Fiedler coordinate (second Laplacian eigenvector): \emph{all} selectors reach ${\approx}68.8\%$, near broadcast ($70.4\%$) and far above no-federation ($61.2\%$). Once adjacency implies data similarity, any neighbor is as good as another: the topology is clearly informative, but there is no gap between informed and random selection. The method's value instead emerges under \emph{strong} heterogeneity.

\begin{figure}[t]
\centering
    \textcolor{blue}{\includegraphics[width=\linewidth]{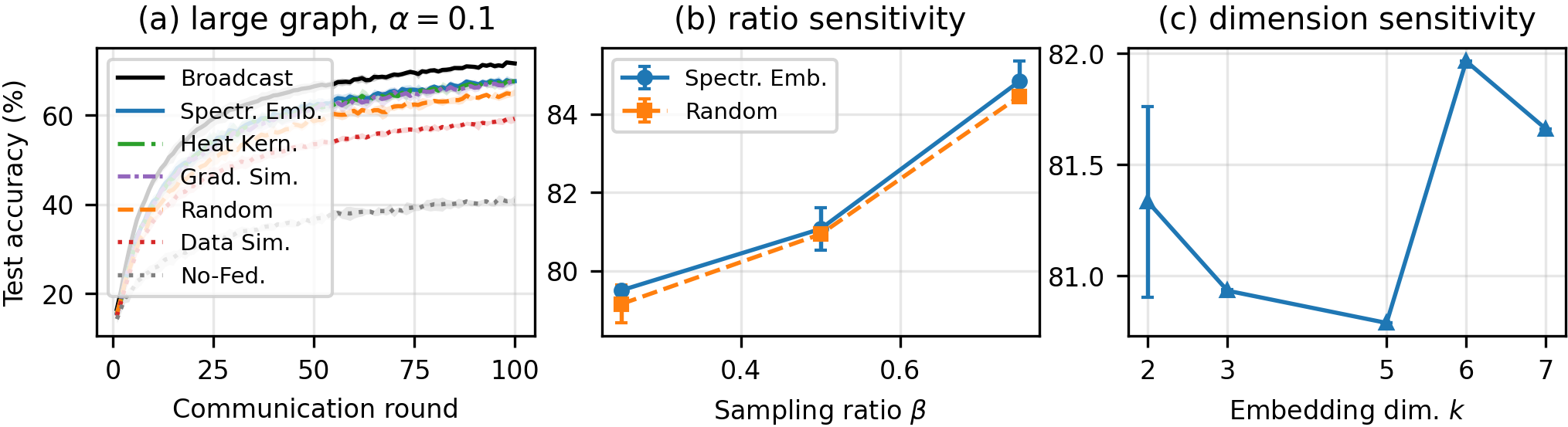}}
    \caption{(a) Convergence on the \textit{large} (Les Mis\'erables) graph (FMNIST, Dirichlet, $\alpha{=}0.1$); all curves are four-seed means ($\pm$std bands). (b,c) Sensitivity of spectral embedding on the \textit{small} graph ($\alpha{=}0.3$) to (b) the ratio $\beta$ and (c) the dimension $k$.}
\label{fig:sens}
\end{figure}

In the second setting ($\alpha{=}0.1$, \textit{large} graph, FMNIST), the advantage \emph{amplifies}: over four seeds, spectral embedding and heat kernel reach $67.5{\pm}0.5\%$ and $67.1{\pm}0.6\%$, outperforming random ($64.5{\pm}0.5\%$) by a ${\sim}3$-point margin (Fig.~\ref{fig:sens}a). The \textit{Data Sim.}\ baseline performs worst ($58.5\%$): aggregating exclusively with label-similar clients restricts diverse gradient exchange and global information propagation, allowing topology-only selection to easily outperform it. As this data is topology-independent, the gain cannot stem from data-topology alignment; it tracks the graph's structural heterogeneity (e.g., cliques, communities), absent from a homogeneous $100$-node Watts-Strogatz graph where selectors match random ($61.3$ vs.\ $61.3\%$). On the \textit{small} graph, spectral selection matches random performance, as having few neighbors limits selection optimization. The method scales smoothly with the sampling ratio $\beta$ and maintains a variance of ${\sim}1\%$ across the embedding dimension $k$ (Fig.~\ref{fig:sens}b,c). It is similarly stable across the heat-kernel diffusion time $t$: varying $t\in\{0.05,0.01,0.1\}$ changes accuracy by ${<}1\%$, with the kernel naturally reducing to adjacency-based uniform selection as $t\!\to\!0$.

\vspace{-0.3cm}
\section{Conclusion} 
\label{sec:concl}
In this letter we proposed a topology-aware client selection strategy for DFL. By leveraging Laplacian spectral embeddings and diffusion kernels, each client samples a structurally relevant subset of its neighbors, outperforming random selection and approaching the accuracy of broadcasting at a fraction of the communication and energy cost, an attractive trade-off for deployments with limited bandwidth and power. Our study shows that this advantage over random is most pronounced under strong data heterogeneity on larger, structurally rich graphs (achieving a ${\sim}3\%$ accuracy gain), while it vanishes to $0\%$ on small or homogeneous topologies where the structure offers little to exploit. Future work will combine the topological criterion with data- and gradient-aware metrics, and extend the framework to dynamic, time-varying topologies.

\bibliographystyle{IEEEtran} \bibliography{bibliography}

\end{document}